\documentclass[sigconf]{acmart}

\renewcommand\footnotetextcopyrightpermission[1]{} 

\usepackage{graphicx}
\usepackage{xcolor}
\usepackage{colortbl}
\usepackage[ruled,vlined,linesnumbered]{algorithm2e}
\usepackage{multirow}
\usepackage{dblfloatfix} 
\usepackage{pifont}
\usepackage{arydshln}

\DeclareMathOperator*{\argmin}{arg\,min}

\definecolor{myc}{cmyk}{1,0,1,0}
\newcommand{\mycolor}[0]{myc}

\newcommand{\tgseventyfive}[0]{\cellcolor{\mycolor!75}}
\newcommand{\tgsixty}[0]{\cellcolor{\mycolor!60}}
\newcommand{\tgfifty}[0]{\cellcolor{\mycolor!50}}
\newcommand{\tgforty}[0]{\cellcolor{\mycolor!40}}

\newcommand{\tgtwentyfive}[0]{\cellcolor{\mycolor!25}}

\newcommand{\tgten}[0]{\cellcolor{\mycolor!10}}
\newcommand{\tgfive}[0]{\cellcolor{\mycolor!5}}

\newcommand\blfootnote[1]{%
  \begingroup
  \renewcommand\thefootnote{}\footnote{#1}%
  \addtocounter{footnote}{-1}%
  \endgroup
}

\AtBeginDocument{%
  \providecommand\BibTeX{{%
    \normalfont B\kern-0.5em{\scshape i\kern-0.25em b}\kern-0.8em\TeX}}}

\acmConference[DAC '22]{ACM Design Automation Conference}{July 2022}{San Francisco, California}%

\begin{document}

\title{\textsc{EcoFusion}: Energy-Aware Adaptive Sensor Fusion for \\ Efficient Autonomous Vehicle Perception}


\author{Arnav Vaibhav Malawade}
\authornote{Both authors contributed equally to this research.}
\author{Trier Mortlock}
\authornotemark[1]
\author{Mohammad Abdullah Al Faruque}
\affiliation{%
  \institution{University of California, Irvine}
  \city{Irvine}
  \state{California}
  \country{USA}
}


\begin{abstract}
Autonomous vehicles use multiple sensors, large deep-learning models, and powerful hardware platforms to perceive the environment and navigate safely. In many contexts, some sensing modalities negatively impact perception while increasing energy consumption. We propose \textbf{\textsc{EcoFusion}}: an energy-aware sensor fusion approach that uses context to adapt the fusion method and reduce energy consumption without affecting perception performance. \textsc{EcoFusion} performs up to \textbf{9.5\%} better at object detection than existing fusion methods with approximately \textbf{60\%} less energy and \textbf{58\%} lower latency on the industry-standard Nvidia Drive PX2 hardware platform. We also propose several context-identification strategies, implement a joint optimization between energy and performance, and present scenario-specific results.
\end{abstract}


\maketitle

\blfootnote{Accepted to be published in the 59\textsuperscript{th} ACM/IEEE Design Automation Conference (DAC 2022).}

\section{Introduction}

Autonomous vehicles (AVs) are expected to improve mobility and road safety dramatically. However, these benefits come with rising energy costs \cite{bradley2015optimization}. AVs require large deep-learning (DL) models to perceive the environment and safely detect and avoid objects. The computational demands of these models significantly increase the hardware requirements of AVs, such that modern AV electrical/electronic (E/E) systems can require between several hundred watts (W) to over 1 kW of power. For example, the Nvidia Drive PX2, used for Tesla Autopilot from 2016-2018, has a Thermal Design Power (TDP) of 250 W \cite{px2tesla}, and modern successors have TDPs ranging from 500 W to 800 W \cite{Abuelsamid2020orin}.
These power demands can also increase the thermal demands on the vehicle's climate-control system.
When combined, these demands can reduce vehicle range by over \textbf{11.5\%} \cite{lin2018architectural}. This impact is especially limiting for electric vehicles due to their limited battery range and long recharge times \cite{vatanparvar2015battery}. Furthermore, many other autonomous systems, including robotics, unmanned aerial vehicles, and sensor networks, operate in energy-constrained environments \cite{beretta2012design,gokhale2021feel,sen2016context}.

\begin{figure}[!t]
    \centering
    \includegraphics[clip, trim=80 5 100 120, width=\linewidth]{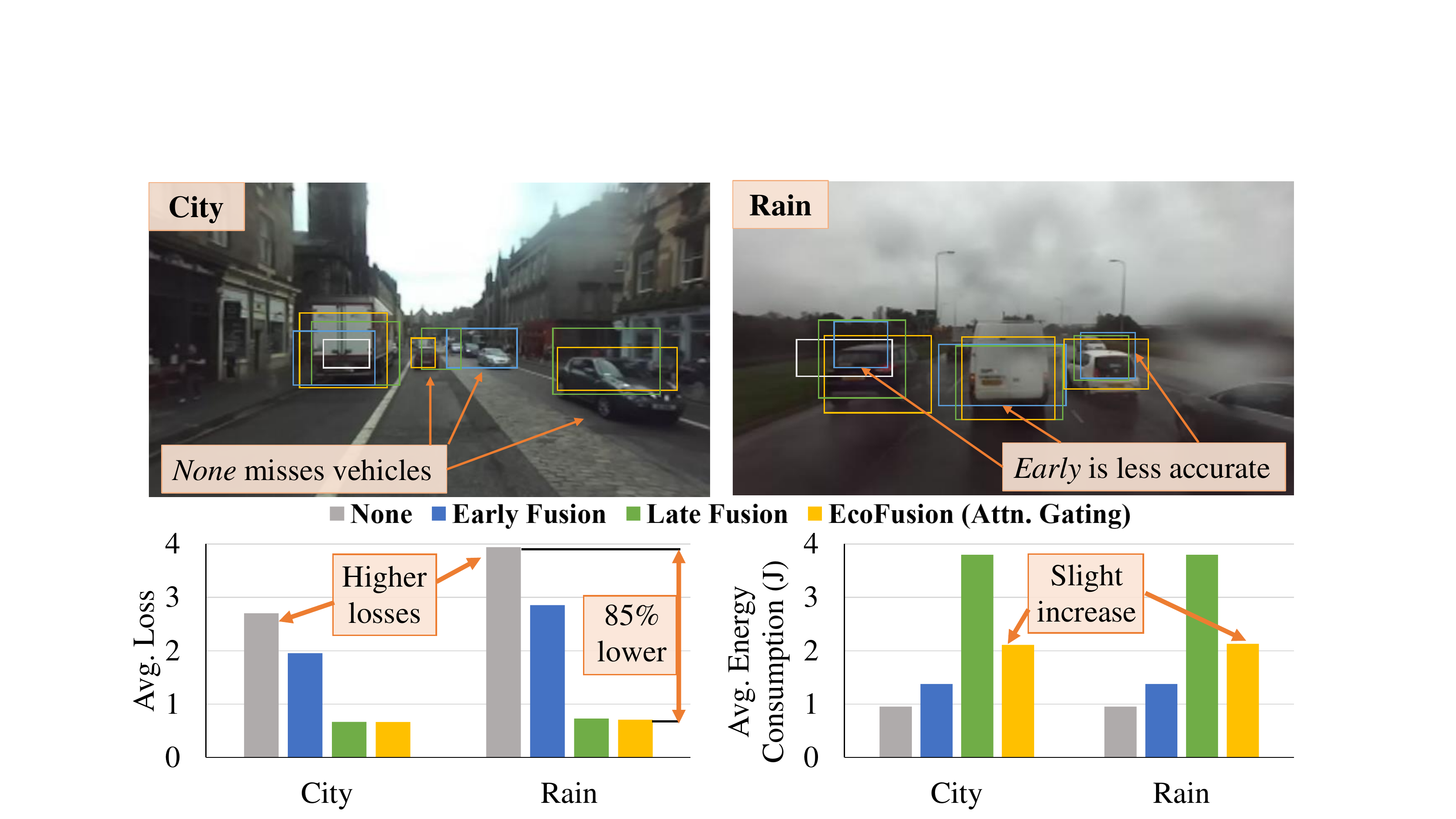}
    \vspace{-6mm}
    \caption{Performance and energy comparison for various AV perception sensor fusion methods in city and rainy driving.}
    \label{fig:mot}
    \vspace{-5mm}
\end{figure}

Recent works have attempted to address the energy demands of AV systems with application-specific hardware design, model pruning, and edge-cloud architectures \cite{pruning, balemans2020resource, lin2018architectural,baidya2020vehicular,malawade2021sage}. These methods have specific downsides as they require expensive hardware modifications, extensive domain knowledge, and consistent network connectivity, respectively. Alternatively, efficient sensor-fusion approaches attempt to combine multiple sensing modalities to achieve good perception performance with less energy than conventional fusion \cite{gokhale2021feel,lee2020accuracy, balemans2020resource}. However, these approaches are also limited because they use statically designed fusion algorithms (e.g., early or late fusion) that can lack robustness in difficult driving scenes~\cite{malawade2022hydrafusion}. Figure~\ref{fig:mot} illustrates the trade-off between performance and energy between different sensor fusion methods for two contexts: city and rain. 
\textit{None} refers to using a single sensor with no fusion, \textit{early fusion} combines raw sensor data before processing, and \textit{late fusion} processes each sensor separately before fusing the final outputs.   
As shown, \textit{no fusion} consumes the least energy but also performs the worst, \textit{late fusion} performs much better but uses almost 3x more energy, and \textit{early fusion} is energy efficient but performs poorly in difficult driving scenarios. 

In summary, our key research challenges include: (i) perceiving the environment accurately in difficult contexts, (ii) reducing the energy consumption of AV perception systems, and (iii) adapting the perception model to the current context to minimize energy consumption without compromising perception performance. We propose \textsc{EcoFusion}: an energy-aware sensor fusion approach that uses context to dynamically switch between different sensor combinations and fusion locations. Our approach can reduce energy consumption without degrading perception performance in comparison to both early and late fusion methods. 
As shown in Figure~\ref{fig:mot}, our approach (shown in gold) achieves higher performance than other fusion methods while significantly reducing energy consumption.

The key contributions of this paper are as follows:
\begin{enumerate}
    \item We propose an energy-aware sensor fusion approach that uses context to adapt the fusion method and reduce energy consumption without affecting perception performance.
    \item We propose novel gating strategies that can identify the context and use it to dynamically adjust the model architecture as part of a joint optimization between energy consumption and model performance.
    \item We benchmark the hardware performance of our approach on the industry-standard Nvidia Drive PX2 autonomous driving platform.
    \item We present an in-depth analysis of the performance of each sensing modality in a range of difficult driving contexts.
\end{enumerate}

\section{Related Work}
\label{sec:relatedwork}
In past years, research on energy-efficient AVs has focused mainly on reducing the energy needs for locomotion and actuation. However, due to the rise in DL perception algorithms and the computational requirements of modern AVs, minimizing the energy consumption of AV E/E systems is becoming a core problem \cite{baxter2018review,bradley2015optimization}. Authors in \cite{balemans2020resource} focus on improving computational efficiency through algorithmic changes for a camera-lidar AV platform while using knowledge-based network pruning in their DL model. 
Selectively fusing sensors, as done in \cite{chen2019selective}, also has potential benefits to save computational energy on AVs.
Distinct from these methods, our approach utilizes the context of the environment to enable further energy optimization for AVs.
Studies have demonstrated the value of context identification, such as in \cite{lee2020accuracy}, where authors propose altering the power levels and operating state of an AV lidar sensor depending on the environmental factors, such as the vehicle's speed, to improve perception efficiency. Likewise, \cite{gokhale2021feel} proposes adjusting the sensing frequency for indoor robot localization according to environmental dynamics. 
However, these approaches are limited as they rely on statically designed context-based rules, whereas our approach employs a self-adaptive design to learn the context of the environment dynamically.  

Trade-offs between the energy and performance of deep neural networks (DNNs), like those used in AV perception, have also been studied. 
\cite{mullapudi2018hydranets} improves the computational efficiency of DNNs for classification by using component-specialization during training and component-selection during inference.
\cite{zhang2018exploring} presents a structure simplification procedure that removes redundant neurons within DNNs.
\cite{tann2016runtime} performs incremental training with DNNs to consider energy-accuracy trade-offs at run-time. 
Unlike our approach, these works are only applied to classification using a single input modality and do not incorporate context. Additionally, we tackle the complex, cross-domain problem of AV energy optimization with our dynamic sensor fusion architecture, and present experiments involving real AV hardware. 

\section{Problem Formulation}
\label{sec:formulation}
Here we detail the formulation for AV object detection and the joint energy-performance optimization implemented in our work.

\subsection{Sensor Fusion for Object Detection}
For each input sample, the goal of an object detector $\phi$ is to utilize the set of sensor measurements in the sample, $\mathbf{X}$, to accurately detect the objects in the scene, $\mathbf{Y}$:
\begin{equation}
    \mathbf{Y} = \phi(\mathbf{X}), \; \text{where} \;
    \mathbf{Y} = \{\mathbf{Y}_{class}^i,  \mathbf{Y}_{reg}^i\}_{i=1 \dots d}  
\end{equation}
where $d$ is the number of objects in the sample. 
$\phi$ can be implemented via conventional sensor fusion techniques, an ML/DL model, or an ensemble of ML/DL models. 
The targets for object $i$ in the sample are defined as follows:
\begin{equation}
    \mathbf{Y}_{class}^i \in \{c_1, c_2, c_3, \dots\}, \;   
    \mathbf{Y}_{reg}^i = [\mu_1 , \nu_1 , \mu_2, \nu_2 ] \in \mathbb{R}^2
\end{equation}
where $\mathbf{Y}_{class}^i$ represents the class of the object (e.g., $c_1$: car, $c_2$: truck, $c_3$: pedestrian) from a set of defined object classes, and $\mathbf{Y}_{reg}^i$ represents the 2D bounding box coordinates of the object in reference to the coordinate frame of the sample. We denote the model's estimate of $\mathbf{Y}$ as $\mathbf{\hat{Y}}$.

Since $\mathbf{X}$ represents data from multiple heterogeneous sensing modalities, sensor fusion can be used to fuse the data to provide a better estimate of $\mathbf{Y}$. In early fusion, the raw sensor inputs are fused before being passed through the object detector as follows:
\begin{equation}
    \mathbf{\hat{Y}}  = \phi(\psi(\mathbf{X}_1, \mathbf{X}_2, \dots , \mathbf{X}_s ))
\end{equation}
where $\psi$ represents the function for fusing the different inputs. In contrast, \textit{late fusion}, involves fusing the outputs of an ensemble of sensor-specific object detectors as follows:
\begin{equation}
    \mathbf{\hat{Y}}_1,\, \mathbf{\hat{Y}}_2,\, \dots, \mathbf{\hat{Y}}_s = \phi_1(\mathbf{X}_1),\,  \phi_2(\mathbf{X}_2),\, \dots , \, \phi_s(\mathbf{X}_s)
\end{equation}
\vspace{-3mm}
\begin{equation}
    \mathbf{\hat{Y}} = \phi(\mathbf{\hat{Y}}_1, \mathbf{\hat{Y}}_2, \dots , \mathbf{\hat{Y}}_s)
\end{equation}
\vspace{-3mm}

\subsection{Energy Modeling}
\begin{figure}[ht]
    \centering
    \includegraphics[clip, trim=150 50 35 238, width=\linewidth]{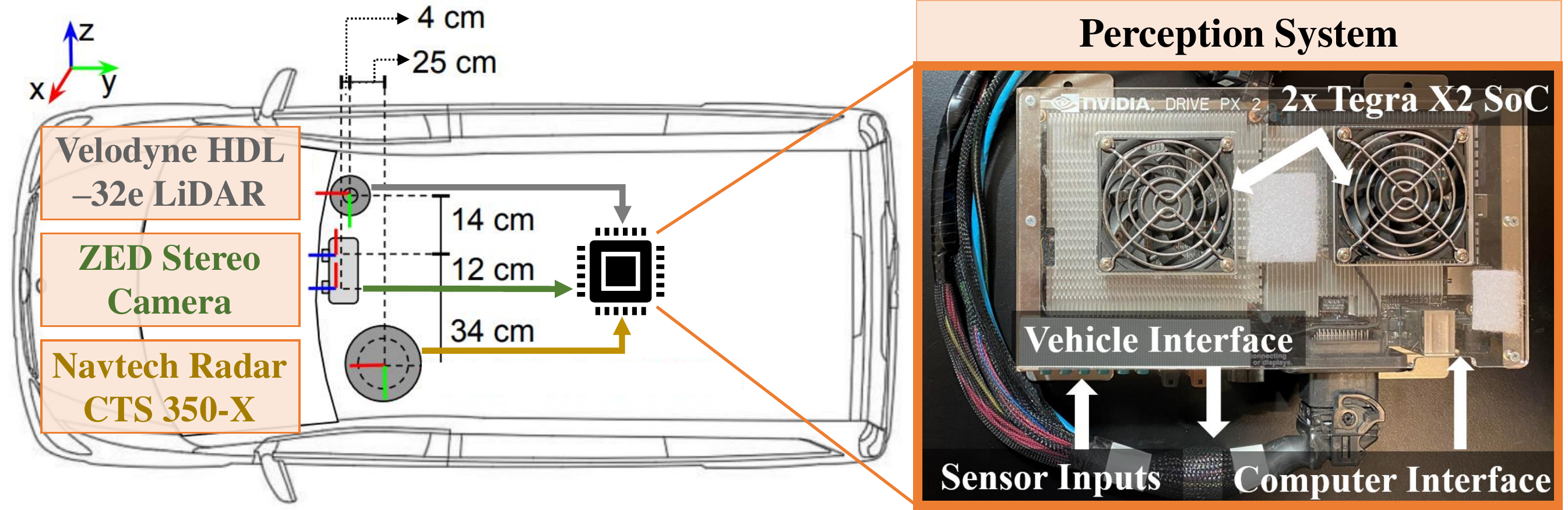}
    \caption{Sensor diagram \cite{sheeny2020radiate} with our Nvidia Drive PX2.}
    \label{fig:hardware}
     \vspace{-5mm}
\end{figure}

In this work, we aim to jointly optimize the energy consumption and performance of the perception system of an AV. To enable this optimization, we use real-world measurements from three different sensors to model the energy consumption of various object detectors $\phi$ on the industry-standard Nvidia Drive PX2 autonomous driving hardware platform, depicted in Figure~\ref{fig:hardware}.
For a given object detector implementation $\phi$ and fixed-size input $\mathbf{X}$, we model energy consumption $\mathbf{E}$ as follows:
\begin{equation}
    \mathbf{E}(\phi, \mathbf{X}) = \mathbf{P(\phi, \mathbf{X})} * t(\phi, \mathbf{X})
\end{equation}
where $t(\phi, \mathbf{X})$ represents the processing latency in seconds, and $\mathbf{P(\phi, \mathbf{X})}$ represents the hardware power consumption in Watts of running input $\mathbf{X}$ through $\phi$ as measured on the hardware. We measured the PX2's average power consumption under load as 45.4 Watts.
Assuming X has a fixed size, we calculate $E(\phi)$ for all $\phi \in \Phi$ offline. Next, we use this energy calculation within a joint optimization framework.
\vspace{-2mm}


\subsection{Joint Energy-Performance Optimization}
We formulate our optimization as a joint minimization problem between energy consumption and model loss. We denote the list of all object detector configurations as $\Phi$.
For each configuration $\phi$ in $\Phi$, we use a model to predict the loss after the outputs of $\phi$ are fused via late fusion, denoted $L_f(\phi)$. The loss is defined as the combined regression and classification loss (using smooth L1 loss and cross-entropy loss, respectively) between the ground-truth $\mathbf{Y}$ and the $\mathbf{\hat{Y}}$ predicted by the model as defined in \cite{ren2015faster}. 
Then, the minimum fusion loss configuration $\phi'$ is identified. We also define the function $\rho$, which determines the set of $\phi$s that have a fusion loss within $\gamma$ of $\phi'$.
This set $\Phi^*$ is defined as follows:
\begin{equation}
    \Phi^* = \rho(L_f(\Phi), \gamma) = \{\phi \in \Phi \; \text{s.t.} \; L_f(\phi) - L_f(\phi') \leq L_f(\phi') + \gamma\}
\end{equation}
where $\gamma$ is the maximum allowable difference in loss between any $\phi$ and $\phi'$ in order for $\phi$ to be included in $\Phi^*$. $\gamma$ can be defined based on the problem and represents the maximum deviation in performance from the best performing configuration $\phi'$ that is allowed to enable the exploration of more efficient configurations. In some cases, maximum performance may not be necessary, so energy can be saved by increasing $\gamma$. Otherwise, if maximum performance is desired, then $\gamma$ can be set to 0, so only $\phi'$ is in $\Phi^*$. 

Given that $E(\phi)$ is known, we have the following joint loss function for each $\phi$ in $\Phi^*$:
\begin{equation}
    L_{joint}(\phi, \lambda_{E}) = (1-\lambda_{E}) * L_f(\phi) + \lambda_{E} * E(\phi)
\end{equation}
where $L(\phi)$ and $E(\phi)$ represent the predicted fusion loss and energy consumption, respectively, of $\phi$; and $\lambda_{E} \in [0.0$ -- $1.0]$ is the weighting factor that weights the importance of energy consumption vs. performance in the joint optimization. Next, we select $\phi^*$, a configuration in $\Phi^*$ which lies on the Pareto frontier of the following minimization:
\begin{equation}
\label{opt}
    \phi^* = \argmin_{\forall \phi \in \Phi^*}(L_{joint}(\phi, \lambda_{E}))
\end{equation}
After $\phi^*$ is identified, it is executed to produce the final set of detections $\mathbf{\hat{Y}}$.


\section{\textsc{EcoFusion} Methodology}
\label{methodology}
\begin{figure}[!ht]
    \centering
    \includegraphics[clip, trim= 15 366 502 10, width=\linewidth]{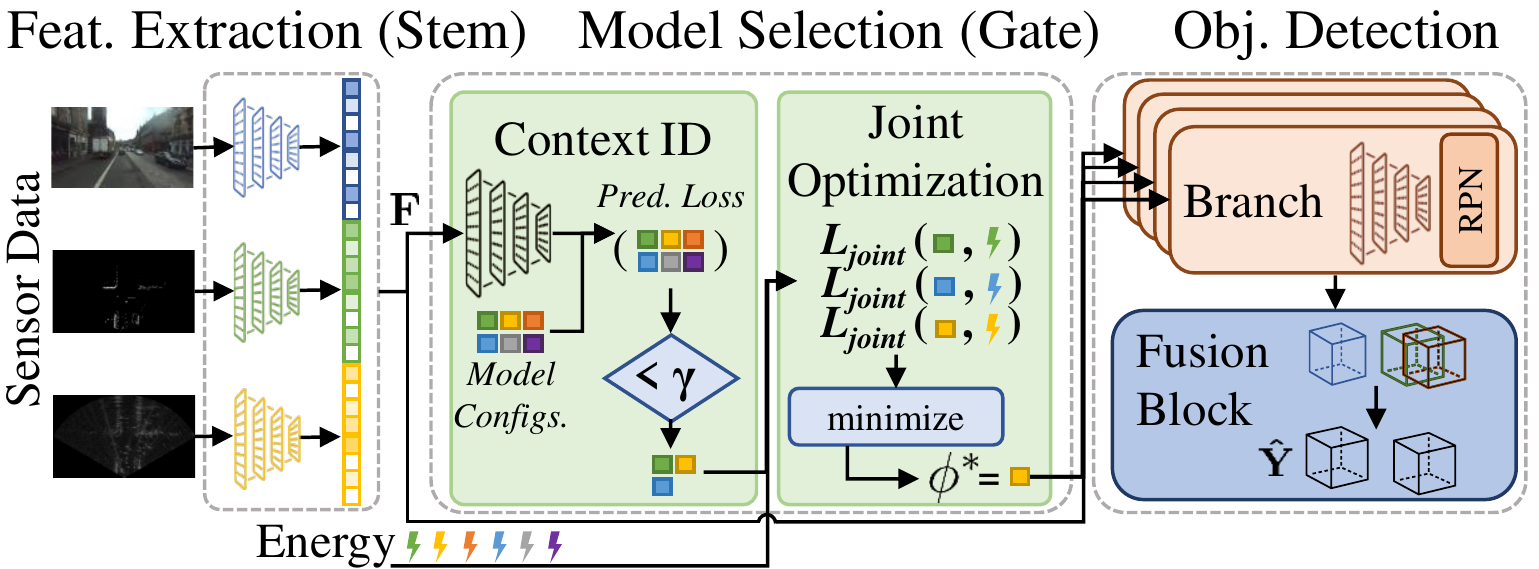}
    \caption{Our proposed \textsc{EcoFusion} framework.}
    \label{fig:archi}
     \vspace{-4mm}
\end{figure}

We propose \textsc{EcoFusion}, a novel adaptive sensor fusion approach that jointly optimizes performance and energy consumption by identifying the context of an environment before subsequently adapting the model and fusion architecture. Our model can: (i) adapt between using no fusion, early fusion, and late fusion, (ii) select from one or more radar, lidar, or camera sensor inputs, and (iii) execute different types of fusion simultaneously depending on what it determines is the best execution path to minimize loss and energy consumption in the current context jointly.

The workflow for our approach is shown in Figure~\ref{fig:archi} and is detailed in Algorithm \ref{alg:ecofusion}. 
First, sensor measurements are passed through modality-specific stem models, which produce an initial set of features $\mathbf{F}$ for each sensor. Next, the gate model uses $\mathbf{F}$ and the set of possible model configurations $\Phi$ to estimate the loss of each possible configuration for the given inputs. After selecting the candidates for optimization using $\gamma$, we pass these candidates $\Phi^*$, their known energy consumption $E$, and their estimated losses $L_f$ to produce $L_{joint}$ for the optimization function. Then, the $\phi$ with the lowest $L_{joint}$, denoted $\phi^*$, is selected to execute as is done in Equation \ref{opt}. Since each $\phi$ represents an ensemble of one or more object detectors, denoted as branches, we run each branch in $\phi^*$ with its expected inputs and collect the results $\mathbf{\hat{Y}}^*$. These are then fused using our late fusion block, producing a final set of detections $\mathbf{\hat{Y}}$. The following subsections elaborate on the different components in our approach.
\begin{algorithm}
\caption{\textsc{EcoFusion} Algorithm}
\DontPrintSemicolon
\label{alg:ecofusion}
\KwIn{$\mathbf{X}$, $\lambda_E$, $\Phi$, $\gamma$, $E(\Phi)$}
\KwOut{Object Detections $\mathbf{\hat{Y}}$}

Initialize feature vector $\mathbf{F}$ and branch output vector $\mathbf{\hat{Y}}^*$.

\For {s in sensors}{
$\mathbf{F}[s] \gets stem(s)$ \tcp*{extract features by modality}
}
$L_f(\Phi) \gets$ gate$(\mathbf{F},\Phi)$ \tcp*{estimate model losses}
$\Phi^* \gets \rho(L_f(\Phi), \gamma)$ \tcp*{select candidates}

\For{$\phi$ in $\Phi^*$}{
$L_{joint}(\phi,\lambda_{E}) \gets (1-\lambda_{E}) * L_f(\phi) + \lambda_{E} * E(\phi)$
}
$\phi^* \gets \argmin_{\forall \phi \in \Phi^*}(L_{joint}(\phi,\lambda_{E}))$ \tcp*{joint opt.}

\For {$\text{branch}$ in $\phi^*$}{
$\mathbf{\hat{Y}}^*[branch] \gets branch(\mathbf{F^*})$\tcp*{pass subset of $\mathbf{F}$}
}

$\mathbf{\hat{Y}} \gets fusion\_block(\mathbf{\hat{Y}}^*)$ \tcp*{fuse branch detections}

\end{algorithm}
%
%
%
\subsection{Stem Model}
The stem models are implemented as a small set of CNN layers that produce an initial set of features for each input modality. The stems are modality-specific, so there is one stem for each type of sensor used. The collection of features $F$ output by the stems is collectively passed to the gate model to identify the context and select the set of branches to execute. Then, $F$ is input to the selected branches.

\subsection{Context-Aware Gating Model}
We implement several gating strategies to estimate the fusion losses of each sensor configuration and facilitate the selection of $\phi^*$. The goal of each gating model is to (i) identify the context based on the input features, (ii) estimate the performance of each model configuration in the context, and (iii) compute the optimization result and use it to select  $\phi^*$. Next, we detail the different methods we implemented for performing steps (i) and (ii).

\subsubsection{Knowledge Gating}
Our Knowledge Gating approach uses domain knowledge on the performance of each modality in different driving conditions to statically decide the best configuration for each rigidly-defined driving context (e.g., rain, snow, city, motorway). This gating approach assumes the context can be identified from external sources, such as weather information, GPS location, and time of day. Also, it assumes that the set of possible contexts is finite, which may limit scalability.

\subsubsection{Deep Gating}
This approach uses a deep-learning model with three CNN layers and one MLP layer to predict the loss for each model configuration for a given set of inputs. Then, the optimization function is run on these outputs.

\subsubsection{Attention Gating}
This approach is identical to the Deep Gating model, except for the addition of a self-attention layer to enable the gate to identify important areas of the input feature map.

\subsubsection{Loss-Based Gating}
In this strategy, the \textit{a posteriori} ground-truth loss from each configuration for a given input is used to select $\phi^*$. Thus, this implementation is not deployable in the real world but represents the theoretical best-case performance for a gate model that can perfectly predict the fusion loss of every configuration for every input. 

\subsection{Branch Models}
The branches in the model take the form of various object detectors.
Each branch performs object detection by implementing a Faster R-CNN \cite{ren2015faster} object detector containing a ResNet-18 CNN model \cite{he2016deep} to extract features from input images and a Region Proposal Network (RPN) to propose object locations across the feature map.
The RPN proposals are then fed through a region-of-interest layer that predicts $Y_{class}^i$,  $Y_{reg}^i$ for each box $i$, as well as the confidence scores for the predicted boxes. 
We split each ResNet-18 model after the first convolution block, such that the first block becomes the stem, and the remaining three convolution blocks are used in each branch.
Each branch can be configured to process either a single sensor or a set of sensors. In this work, we implement one branch for each input sensor and three early fusion branches that fuse both homogeneous and heterogeneous sets of sensors. Using the gate to select the branches, our model can dynamically choose between no fusion, early fusion, late fusion, and combinations of the three. 

\subsection{Fusion Block}
The fusion block is implemented via a typical late-fusion algorithm. The detections from any number of branches are first converted to a uniform coordinate system before being statistically processed and fused using the weighted box fusion method from \cite{solovyev2021weighted}. This process helps refine the accuracy of the bounding box predictions by reinforcing predictions with high confidence and overlap with other predictions.  

\section{Experiments}
\label{exp}
In our experiments, we used the RADIATE \cite{sheeny2020radiate} dataset, which provides annotated real-world object detection data from an AV with the following sensors: a Navtech CTS350-X radar, a Velodyne HDL-32e lidar, and a ZED stereo camera. The following classes of objects are annotated in the dataset: \{\textit{car, van, truck, bus, motorbike, bicycle, pedestrian, group of pedestrians}\}. The dataset consists of various difficult driving contexts (\textit{e.g.}, \textit{rain, fog, snow, city, motorway}) that are challenging for typical object detectors. In \textsc{EcoFusion}, we use a 70:30 train-test split across the dataset and train our model with all of the stems and branches enabled using supervised learning. Next, we take the trained stem and branch outputs and use them to separately train the gate model to select the branches that produce the lowest loss for a given stem output ($\mathbf{F}$).
We evaluate each model's performance at object detection using average loss and mean average precision (mAP), which is widely used for benchmarking object detection models \cite{ren2015faster, everingham2010pascal}. We compute the mAP for bounding boxes with an intersection-over-union (IoU) $\geq0.5$, aligning with the PASCAL Visual Object Classes (VOC) Challenge \cite{everingham2010pascal}. 
We calculated the energy consumption of each model configuration $\phi \in \Phi$ on the Nvidia Drive PX2 shown in Figure~\ref{fig:hardware}.
We ignore the energy consumed by the gate models as we measured that they have negligible energy consumption ($<0.005$~J) compared to the stems and branches of the model after TensorRT compilation.
In all of our experiments, we set $\gamma=0.5$ as we experimentally determined that it ensures performance at least as good as early and late fusion while enabling energy optimization. However, we note that $\gamma$ can be tuned based on the requirements for a given application.

\vspace{-3mm}
\begin{figure}[htb]
    \centering
    \includegraphics[clip, trim=135 10 105 10, width=\linewidth]{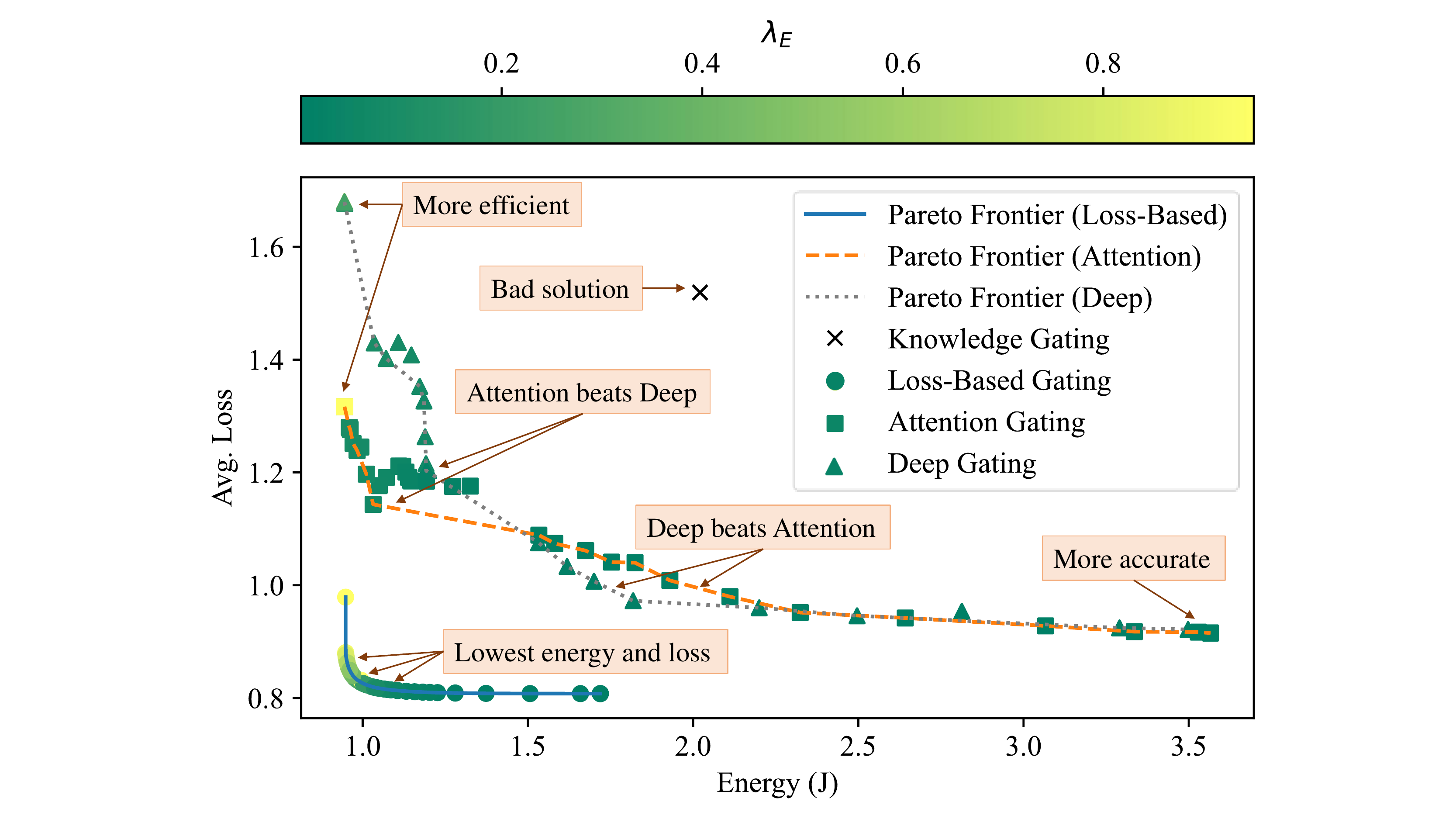}
    \vspace{-5mm}
    \caption{Analysis of the energy-loss trade-off of \textsc{EcoFusion}'s optimization function with gating models and $\lambda_E$ values.}
    \label{fig:optimization}
     \vspace{-5mm}
\end{figure}

\subsection{Joint Optimization Analysis}
We evaluated the trade-off between the performance (model loss) and energy consumption (in Joules) for each gating model in Figure~\ref{fig:optimization}. We varied $\lambda_E$ between 0-1.0, where each point in the chart is color-coded according to its $\lambda_E$ value. As shown, tuning $\lambda_E$ higher or lower skews the model towards either increasing energy efficiency or increasing performance, respectively, so $\lambda_E$ should be chosen depending on the requirements for a given application. The configuration for \textit{Loss-Based} that best minimizes both objectives is $\lambda_E=0.5$ with a loss of 0.966 and energy consumption of 0.844~J. \textit{Attention} and \textit{Deep} have similar Pareto frontiers, but \textit{Attention} achieves better solutions for higher $\lambda_E$ values while \textit{Deep} achieves slightly lower loss with some low $\lambda_E$ values. The gap between \textit{Attention/Deep} and \textit{Loss-Based} is likely due to modeling limitations and could potentially be closed using larger or more advanced gate models.
For \textit{Attention}, $\lambda_E=1$ (most energy efficient) results in a loss of 1.317 and an energy consumption of 0.945~J, while $\lambda_E=0$ (best performing) results in a loss of 0.9153 and an energy consumption of 3.566~J. As shown by the nearly flat trend on the right side of the plot, \textit{Deep} and \textit{Attention} can reduce energy significantly with little effect on loss by tuning $\lambda_E$.
\textit{Knowledge} is statically programmed such that, for each scenario type, we use domain knowledge to manually select the best sensor combination to use. Due to these constraints, \textit{Knowledge} can be less efficient in some scenarios and is not tunable with our optimization.

\subsection{Energy and Performance Evaluation}
Our results for energy consumption and performance evaluation are shown in Table~\ref{tab:energy_perf_eval}. In all of our experiments, early fusion takes in both cameras and lidar as input, while late fusion uses both cameras, lidar, and radar. 
The energy consumption and latency increase as the fusion method is varied from none to early to late, which is as expected as the latter methods require increasingly larger detection pipelines. The single-sensors are the most efficient, but their mAP scores vary widely from 67\% to 79\%, likely due to inconsistent performance across scenarios. Early fusion is faster, more efficient, and achieves a higher mAP score and than late fusion; however, early fusion is insufficiently robust in poor driving conditions as will be discussed in Section \ref{subsec:scenario-eval}. \textsc{EcoFusion} with $\lambda_E=0.01$ achieves higher mAP than all other methods with less energy than late fusion. With $\lambda_E=0.05$, \textsc{EcoFusion} still outperforms early fusion with less energy usage. As stated in \cite{lin2018architectural}, an AV must be able to process inputs at least once every 100 ms (10 frames per second) to ensure safety. In addition to meeting this latency requirement, \textsc{EcoFusion} also executes faster than both early and late fusion, which can improve safety and responsiveness by enabling the AV to process inputs more frequently.
With $\lambda_E = 0.01$, \textsc{EcoFusion} achieves a mAP score \textbf{5.1\%} and \textbf{9.5\%} higher than early and late fusion, respectively, with \textbf{60\%} less energy and \textbf{58\%} lower latency than late fusion. 

\begin{table}[htb]
    \centering
    \begin{tabular}{p{40pt} p{60pt} p{30pt} p{30pt} p{30pt}}
    \hline
Fusion Type & Configuration & mAP (\%) & Energy (J) & Latency (ms)\\\hline
\multirow{4}{30pt}{None} & L. Camera ($C_L$) & 74.48\% & 0.945 & 21.57\\
& R. Camera ($C_R$) & 79.00\% & 0.945 & 21.57\\
& Radar ($R$) & 67.74\% & 0.954 & 21.85\\
& Lidar ($L$) & 70.45\% & 0.954 & 21.85\\\hline
\multirow{1}{30pt}{Early} 
& $C_L + C_R + L$ & 80.26\% & 1.379 & 31.36\\\hline
\multirow{1}{30pt}{Late} 
& $C_L + C_R + L + R$ & 77.98\% & 3.798 & 84.32\\\hline
\multirow{3}{30pt}{\textbf{\textsc{EcoFusion} (Ours)}} &$\lambda_E=0$ & 82.92\% & 3.566 & 81.49\\
&\textbf{$\lambda_E=0.01$} & \textbf{84.32\%} & \textbf{1.533} & \textbf{35.14}\\
&\textbf{$\lambda_E=0.05$} & \textbf{82.16\%} & \textbf{1.110} & \textbf{25.43}\\\hline
    \end{tabular}
    \caption{Energy Consumption and Performance Evaluation}
    \label{tab:energy_perf_eval}
    \vspace{-8mm}
\end{table}

\subsection{Gating Method Evaluation}
Table~\ref{tab:gating_eval} shows mAP, loss, and energy results from evaluating our gating strategies at different $\lambda_E$ values. With $\lambda_E = 0$, the models tend to pick better-performing branches regardless of their energy consumption. As $\lambda_E$ increases, the joint optimization significantly reduces energy consumption while keeping loss within $\gamma$ of the lowest-loss configuration. Although \textit{Knowledge} achieves decent mAP scores, it lacks tunability and thus achieves the same loss and energy consumption for all $\lambda_E$; the encoded knowledge would need to be manually updated to adjust the trade-off. \textit{Loss-Based} achieves the lowest loss and energy consumption but a lower mAP than \textit{Deep} and \textit{Attention}. This result is likely because loss is not perfectly correlated with mAP score; mAP primarily scores object classification over properly aligned bounding boxes, while loss is measured across both classification and box regression.
Overall, \textit{Attention} performs slightly better than \textit{Deep} and offers the best trade-off of performance and energy.

\begin{table}[htb]
    \centering
    \begin{tabular}{c c c c c}
    \hline
$\lambda_E$ &Gating Method & mAP (\%) & Avg. Loss & Energy (J)\\\hline
0 & \textit{Knowledge} & 82.43\% & 1.519 & \textbf{2.021}\\
0 & \textit{Deep} & 82.68\% & \textbf{0.915} & 3.556\\
0 & \textit{Attention} & \textbf{82.92}\% & \textbf{0.915} & 3.566\\\hdashline
0 & \textit{Loss-Based} & 82.50\% & 0.808 & 1.719\\\hline
0.01 & \textit{Knowledge} & 82.43\% & 1.519 & 2.021\\
0.01 & \textit{Deep} & 83.72\% & 1.124 & \textbf{1.457}\\
0.01 & \textit{Attention} & \textbf{84.32\%} & \textbf{1.089} & 1.533\\\hdashline
0.01 & \textit{Loss-Based} & 81.65\% & 0.809 & 1.280\\\hline
0.1 & \textit{Knowledge} & \textbf{82.43\%} & 1.519 & 2.021\\
0.1 & \textit{Deep} & 81.98\% & 1.432 & 1.008\\
0.1 & \textit{Attention} & 79.72\% & \textbf{1.280} & \textbf{0.960}\\\hdashline
0.1 & \textit{Loss-Based} & 79.70\% & 0.818 & 1.044\\\hline
    \end{tabular}
    \caption{Gating method evaluation.}
    \label{tab:gating_eval}
     \vspace{-6mm}
\end{table}

\subsection{Scenario-Specific Evaluation}
\label{subsec:scenario-eval}
\begin{figure*}[htb]
    \centering
    \includegraphics[clip, trim=5 368 190 10, width=\textwidth]{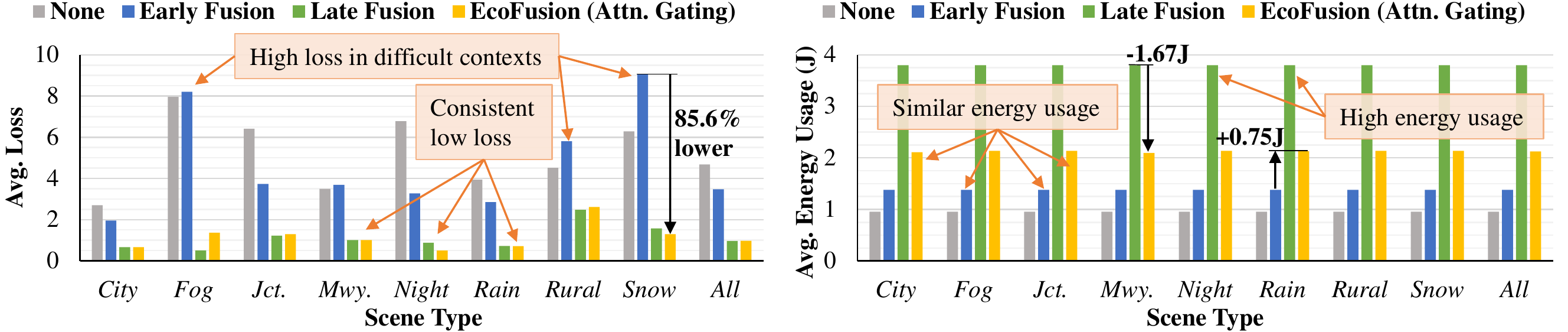}
    \vspace{-5mm}
    \caption{Average loss and energy consumption per scenario for each fusion method. Junction and Motorway are abbreviated as Jct. and Mwy., respectively. \textsc{EcoFusion} achieves low loss across scenes with 43.7\% lower energy consumption than late fusion.}
    \label{fig:scene_eval}
\end{figure*}

Figure~\ref{fig:scene_eval} shows loss and energy results for different driving scenarios in the dataset. We evaluated no fusion (radar-only), early fusion, late fusion, and \textsc{EcoFusion} with \textit{Attention Gating}. As shown in the figure, \textsc{EcoFusion} performs similarly to late fusion in terms of loss across all scenarios. It is also clear that early fusion performs poorly in the difficult driving conditions present in the \textit{Fog} and \textit{Snow} scenarios. Late fusion is more robust and achieves relatively good performance across scenes; however, late fusion also consumes significantly more energy than all other methods. In contrast, \textsc{EcoFusion}'s energy efficiency is on-par with early fusion and is significantly lower than that of late fusion. No fusion was the most energy-efficient but also had the highest overall loss.

\subsection{Discussion}
\label{discussion}

\subsubsection{Practicality}
Since we evaluated our approach with the industry-standard Nvidia Drive PX2 autonomous driving platform, it is clear that our approach can save energy on real-world AV hardware while meeting real-time latency constraints. Furthermore, by achieving better object detection performance with lower latency, our approach improves safety and robustness over existing methods. Our evaluation on a diverse driving dataset proves that our approach is robust across scenarios and is thus more practical for real-world driving. To implement \textsc{EcoFusion} on a real driving system, the designer would first need to train the model on the appropriate dataset before selecting the best $\lambda_E$ and $\gamma$ for their design requirements. Then, the model can be compiled for hardware using TensorRT or a similar library and integrated into the AV stack. 

\subsubsection{Sensor Clock Gating}
\begin{table*}[htb]
    \centering
    \begin{tabular}{c c c c c c c c c c}
    \hline
        \multirow{2}{*}{Fusion Method} & \multicolumn{9}{c}{Avg. Energy Consumption (J) by Scene Type}\\\cline{2-10}
        & \textit{City} & \textit{Fog} & \textit{Jct.} & \textit{Mwy.} & \textit{Night} & \textit{Rain} & \textit{Rural} & \textit{Snow} & \textit{Overall}\\\hline
        Late Fusion & \tgten 13.27 & \tgten{13.27} & \tgten13.27 & \tgten13.27 & \tgten13.27 & \tgten13.27 & \tgten13.27 & \tgten13.27 & \tgten13.27 \\
        \textsc{EcoFusion} (Ours) &\tgfifty\textbf{5.45} & \tgfive13.96 & \tgseventyfive\textbf{2.87} & \tgseventyfive\textbf{2.87} & \tgtwentyfive\textbf{12.10} & \tgfive13.29 & \tgsixty\textbf{3.81} & \tgfive13.96 & \tgforty\textbf{6.45}\\\hline
        \textbf{\textsc{EcoFusion} Energy Savings} & \tgfifty\textbf{58.91\%} & \tgfive-5.15\% & \tgseventyfive\textbf{78.40\%} & \tgseventyfive\textbf{78.40\%} & \tgtwentyfive\textbf{8.81\%} & \tgfive-0.09\% & \tgsixty\textbf{71.28\%} & \tgfive-5.15\% & \tgforty\textbf{51.41\%}\\\hline
    \end{tabular}
    \caption{Combined sensor and AV hardware platform energy consumption in each driving scenario.}
    \label{tab:power_gating}
     \vspace{-5mm}
\end{table*}
More energy could be saved by disabling unused sensors using clock gating. The Navtech CTS350-X radar uses 24~W \cite{navtechdatasheet}, the Velodyne HDL-32E lidar uses 12~W \cite{velodynedatasheet} and the ZED camera uses 1.9~W \cite{zeddatasheet}, so reducing sensor energy usage can significantly improve AV efficiency. 
Temporal modeling can enable the context to be estimated across time instead of for a single input, allowing clock gating for specific periods. In Table~\ref{tab:power_gating}, we analyze the benefits of sensor clock gating with our \textit{Knowledge Gating} approach in each driving scenario since it uses external context to inform sensor selection. We also show baseline results with late fusion across the four sensors.
Using the power consumption $P$ and measurement frequency $f$ of each sensor $s$, we estimate the energy that could be saved by stopping measurements without slowing the motor's rotation. We cannot completely power gate the rotating lidar and radar sensors because they have inertia and require several seconds to get back up to speed from a stand-still, which can compromise safety. 
We model the energy consumption $E_s$ of each sensor and the total energy consumption $E_{total}$ as follows:
\begin{equation}
    E_s = (P_s^{meas.} + P_s^{motor}) * 1/f_s, \;\; P_s^{meas.} = P_s - P_s^{motor}
\end{equation}
\begin{equation}
    E_{total} = E(\phi) + \sum_{s \in \phi}E_s
\end{equation}
where $\phi$ is the model configuration defined for the context. 
After our calculation, we set $P^{meas.} = 0$ to simulate clock gating of the sensor. The Navtech CTS350-X consumes 2.4~W to spin the motor, so its $P^{meas.} = 21.6$~W. 
Based on comparable lidar motor models, we estimate the Velodyne HDL-32E's $P^{meas.}=9.6$~W. 
As shown in Table~\ref{tab:power_gating}, \textsc{EcoFusion} would use up to \textbf{78.40\%} less energy than late fusion in common driving scenarios. Our approach uses slightly more energy than late fusion in more difficult driving scenarios, but these scenarios are rare, so overall energy consumption is still lower. On average, clock gating unused sensors with \textsc{EcoFusion} uses \textbf{51.41\%} less energy than running all sensors with late fusion and \textbf{43.90\%} less energy than \textsc{EcoFusion} without sensor clock gating.


\section{Conclusion}
\label{sec:conclusion}

This paper introduces \textsc{EcoFusion} --- a novel adaptive sensor fusion approach that uses contextual information to adapt its architecture and jointly optimize performance and energy consumption. We show that \textsc{EcoFusion} outperforms early and late fusion in terms of mAP (\textbf{84.32\%} vs. 80.26\% and 77.98\%), with similar energy consumption and latency to early fusion. We also demonstrate that in difficult driving contexts, \textsc{EcoFusion} is more robust than early fusion (up to \textbf{85.6\%} lower loss) and more efficient than late fusion (\textbf{60\%} less energy). We additionally propose and evaluate multiple gating strategies and find that a learned strategy outperforms a knowledge-based strategy. Overall, we show that an energy-aware adaptive sensor fusion approach can significantly improve the energy efficiency and perception performance of AVs. 

\begin{acks}
This work was partially supported by the National Science Foundation (NSF) under awards CMMI-1739503 and CCF-2140154. Any opinions, findings, conclusions, or recommendations expressed in this paper are those of the authors and do not necessarily reflect the views of the funding agencies.
\end{acks}

\bibliographystyle{ACM-Reference-Format}
\bibliography{bibliography}


\begin{thebibliography}{27}


\ifx \showCODEN    \undefined \def \showCODEN     #1{\unskip}     \fi
\ifx \showDOI      \undefined \def \showDOI       #1{#1}\fi
\ifx \showISBNx    \undefined \def \showISBNx     #1{\unskip}     \fi
\ifx \showISBNxiii \undefined \def \showISBNxiii  #1{\unskip}     \fi
\ifx \showISSN     \undefined \def \showISSN      #1{\unskip}     \fi
\ifx \showLCCN     \undefined \def \showLCCN      #1{\unskip}     \fi
\ifx \shownote     \undefined \def \shownote      #1{#1}          \fi
\ifx \showarticletitle \undefined \def \showarticletitle #1{#1}   \fi
\ifx \showURL      \undefined \def \showURL       {\relax}        \fi
\providecommand\bibfield[2]{#2}
\providecommand\bibinfo[2]{#2}
\providecommand\natexlab[1]{#1}
\providecommand\showeprint[2][]{arXiv:#2}

\bibitem[\protect\citeauthoryear{Abuelsamid}{Abuelsamid}{2020}]%
        {Abuelsamid2020orin}
\bibfield{author}{\bibinfo{person}{Sam Abuelsamid}.}
  \bibinfo{year}{2020}\natexlab{}.
\newblock \showarticletitle{{Nvidia Cranks Up And Turns Down Its Drive AGX Orin
  Computers}}.
\newblock \bibinfo{journal}{\emph{Forbes}} (\bibinfo{date}{Jun}
  \bibinfo{year}{2020}).
\newblock
\urldef\tempurl%
\url{https://www.forbes.com/sites/samabuelsamid/2020/05/14/nvidia-cranks-up-and-turns-down-its-drive-agx-orin-computers}
\showURL{%
\tempurl}


\bibitem[\protect\citeauthoryear{Baidya~\textit{et al.}}{Baidya~\textit{et
  al.}}{2020}]%
        {baidya2020vehicular}
\bibfield{author}{\bibinfo{person}{Sabur Baidya~\textit{et al.}}}
  \bibinfo{year}{2020}\natexlab{}.
\newblock \showarticletitle{Vehicular and edge computing for emerging connected
  and autonomous vehicle applications}. In \bibinfo{booktitle}{\emph{DAC '20}}.
  IEEE, \bibinfo{pages}{1--6}.
\newblock


\bibitem[\protect\citeauthoryear{Balemans~\textit{et al.}}{Balemans~\textit{et
  al.}}{2020}]%
        {balemans2020resource}
\bibfield{author}{\bibinfo{person}{Dieter Balemans~\textit{et al.}}}
  \bibinfo{year}{2020}\natexlab{}.
\newblock \showarticletitle{Resource efficient sensor fusion by knowledge-based
  network pruning}.
\newblock \bibinfo{journal}{\emph{Internet of Things}}  \bibinfo{volume}{11}
  (\bibinfo{year}{2020}), \bibinfo{pages}{100231}.
\newblock


\bibitem[\protect\citeauthoryear{Baxter~\textit{et al.}}{Baxter~\textit{et
  al.}}{2018}]%
        {baxter2018review}
\bibfield{author}{\bibinfo{person}{Jared~A Baxter~\textit{et al.}}}
  \bibinfo{year}{2018}\natexlab{}.
\newblock \showarticletitle{Review of electrical architectures and power
  requirements for automated vehicles}. In \bibinfo{booktitle}{\emph{2018
  ITEC}}. IEEE, \bibinfo{pages}{944--949}.
\newblock


\bibitem[\protect\citeauthoryear{Beretta~\textit{et al.}}{Beretta~\textit{et
  al.}}{2012}]%
        {beretta2012design}
\bibfield{author}{\bibinfo{person}{Ivan Beretta~\textit{et al.}}}
  \bibinfo{year}{2012}\natexlab{}.
\newblock \showarticletitle{Design exploration of energy-performance trade-offs
  for wireless sensor networks}. In \bibinfo{booktitle}{\emph{DAC '12}}.
  \bibinfo{pages}{1043--1048}.
\newblock


\bibitem[\protect\citeauthoryear{Bradley~\textit{et al.}}{Bradley~\textit{et
  al.}}{2015}]%
        {bradley2015optimization}
\bibfield{author}{\bibinfo{person}{Justin~M Bradley~\textit{et al.}}}
  \bibinfo{year}{2015}\natexlab{}.
\newblock \showarticletitle{Optimization and control of cyber-physical vehicle
  systems}.
\newblock \bibinfo{journal}{\emph{Sensors}} \bibinfo{volume}{15},
  \bibinfo{number}{9} (\bibinfo{year}{2015}), \bibinfo{pages}{23020--23049}.
\newblock


\bibitem[\protect\citeauthoryear{Chen~\textit{et al.}}{Chen~\textit{et
  al.}}{2019}]%
        {chen2019selective}
\bibfield{author}{\bibinfo{person}{Changhao Chen~\textit{et al.}}}
  \bibinfo{year}{2019}\natexlab{}.
\newblock \showarticletitle{Selective sensor fusion for neural visual-inertial
  odometry}. In \bibinfo{booktitle}{\emph{CVPR '19}}.
  \bibinfo{pages}{10542--10551}.
\newblock


\bibitem[\protect\citeauthoryear{Everingham~\textit{et
  al.}}{Everingham~\textit{et al.}}{2010}]%
        {everingham2010pascal}
\bibfield{author}{\bibinfo{person}{Mark Everingham~\textit{et al.}}}
  \bibinfo{year}{2010}\natexlab{}.
\newblock \showarticletitle{The pascal visual object classes ({VOC})
  challenge}.
\newblock \bibinfo{journal}{\emph{International Journal of Computer Vision}}
  \bibinfo{volume}{88}, \bibinfo{number}{2} (\bibinfo{year}{2010}),
  \bibinfo{pages}{303--338}.
\newblock


\bibitem[\protect\citeauthoryear{Gokhale~\textit{et al.}}{Gokhale~\textit{et
  al.}}{2021}]%
        {gokhale2021feel}
\bibfield{author}{\bibinfo{person}{Vineet Gokhale~\textit{et al.}}}
  \bibinfo{year}{2021}\natexlab{}.
\newblock \showarticletitle{FEEL: fast, energy efficient localization for
  autonomous indoor vehicles}.
\newblock \bibinfo{journal}{\emph{arXiv:2102.00702}} (\bibinfo{year}{2021}).
\newblock


\bibitem[\protect\citeauthoryear{He~\textit{et al.}}{He~\textit{et
  al.}}{2016}]%
        {he2016deep}
\bibfield{author}{\bibinfo{person}{Kaiming He~\textit{et al.}}}
  \bibinfo{year}{2016}\natexlab{}.
\newblock \showarticletitle{Deep residual learning for image recognition}. In
  \bibinfo{booktitle}{\emph{CVPR '16}}. \bibinfo{pages}{770--778}.
\newblock


\bibitem[\protect\citeauthoryear{Lambert}{Lambert}{2016}]%
        {px2tesla}
\bibfield{author}{\bibinfo{person}{Fred Lambert}.}
  \bibinfo{year}{2016}\natexlab{}.
\newblock \bibinfo{title}{{All new Teslas are equipped with NVIDIA's new Drive
  PX 2 AI platform for self-driving - Electrek}}.
\newblock
  \bibinfo{howpublished}{https://electrek.co/2016/10/21/all-new-teslas-are-equipped-with-nvidias-new-drive-px-2-ai-platform-for-self-driving}.
\newblock


\bibitem[\protect\citeauthoryear{Lee~\textit{et al.}}{Lee~\textit{et
  al.}}{2020}]%
        {lee2020accuracy}
\bibfield{author}{\bibinfo{person}{Sanghoon Lee~\textit{et al.}}}
  \bibinfo{year}{2020}\natexlab{}.
\newblock \showarticletitle{Accuracy--power controllable lidar sensor system
  with {3D} object recognition for autonomous vehicle}.
\newblock \bibinfo{journal}{\emph{Sensors}} \bibinfo{volume}{20},
  \bibinfo{number}{19} (\bibinfo{year}{2020}), \bibinfo{pages}{5706}.
\newblock


\bibitem[\protect\citeauthoryear{Lidar}{Lidar}{2021}]%
        {velodynedatasheet}
\bibfield{author}{\bibinfo{person}{Velodyne Lidar}.}
  \bibinfo{year}{2021}\natexlab{}.
\newblock \bibinfo{title}{{Velodyne HDL-32e Datasheet}}.
\newblock
\newblock
\urldef\tempurl%
\url{https://velodynelidar.com/products/hdl-32e/}
\showURL{%
\tempurl}


\bibitem[\protect\citeauthoryear{Lin~\textit{et al.}}{Lin~\textit{et
  al.}}{2018}]%
        {lin2018architectural}
\bibfield{author}{\bibinfo{person}{Shih-Chieh Lin~\textit{et al.}}}
  \bibinfo{year}{2018}\natexlab{}.
\newblock \showarticletitle{The architectural implications of autonomous
  driving: Constraints and acceleration}. In
  \bibinfo{booktitle}{\emph{ASPLOS'18}}. \bibinfo{pages}{751--766}.
\newblock


\bibitem[\protect\citeauthoryear{Malawade, Mortlock, and Faruque}{Malawade
  et~al\mbox{.}}{2022}]%
        {malawade2022hydrafusion}
\bibfield{author}{\bibinfo{person}{Arnav~Vaibhav Malawade},
  \bibinfo{person}{Trier Mortlock}, {and} \bibinfo{person}{Mohammad Abdullah~Al
  Faruque}.} \bibinfo{year}{2022}\natexlab{}.
\newblock \showarticletitle{HydraFusion: Context-Aware Selective Sensor Fusion
  for Robust and Efficient Autonomous Vehicle Perception}. In
  \bibinfo{booktitle}{\emph{ICCPS '22}}. IEEE.
\newblock


\bibitem[\protect\citeauthoryear{Malawade~\textit{et al.}}{Malawade~\textit{et
  al.}}{2021}]%
        {malawade2021sage}
\bibfield{author}{\bibinfo{person}{Arnav Malawade~\textit{et al.}}}
  \bibinfo{year}{2021}\natexlab{}.
\newblock \showarticletitle{SAGE: A Split-Architecture Methodology for
  Efficient End-to-End Autonomous Vehicle Control}.
\newblock \bibinfo{journal}{\emph{ACM TECS}} \bibinfo{volume}{20},
  \bibinfo{number}{5s} (\bibinfo{year}{2021}).
\newblock


\bibitem[\protect\citeauthoryear{Mullapudi~\textit{et
  al.}}{Mullapudi~\textit{et al.}}{2018}]%
        {mullapudi2018hydranets}
\bibfield{author}{\bibinfo{person}{Ravi~Teja Mullapudi~\textit{et al.}}}
  \bibinfo{year}{2018}\natexlab{}.
\newblock \showarticletitle{Hydranets: Specialized dynamic architectures for
  efficient inference}. In \bibinfo{booktitle}{\emph{CVPR '18}}.
  \bibinfo{pages}{8080--8089}.
\newblock


\bibitem[\protect\citeauthoryear{Radar}{Radar}{2021}]%
        {navtechdatasheet}
\bibfield{author}{\bibinfo{person}{Navtech Radar}.}
  \bibinfo{year}{2021}\natexlab{}.
\newblock \bibinfo{title}{{Navtech CTS Series}}.
\newblock
\newblock
\urldef\tempurl%
\url{https://navtechradar.com/clearway-technical-specifications/compact-sensors}
\showURL{%
\tempurl}


\bibitem[\protect\citeauthoryear{Ren~\textit{et al.}}{Ren~\textit{et
  al.}}{2015}]%
        {ren2015faster}
\bibfield{author}{\bibinfo{person}{Shaoqing Ren~\textit{et al.}}}
  \bibinfo{year}{2015}\natexlab{}.
\newblock \showarticletitle{Faster {R-CNN}: Towards real-time object detection
  with region proposal networks}.
\newblock \bibinfo{journal}{\emph{Advances in neural information processing
  systems}}  \bibinfo{volume}{28} (\bibinfo{year}{2015}),
  \bibinfo{pages}{91--99}.
\newblock


\bibitem[\protect\citeauthoryear{{Samal \textit{et al.}}}{{Samal \textit{et
  al.}}}{2020}]%
        {pruning}
\bibfield{author}{\bibinfo{person}{K. {Samal \textit{et al.}}}}
  \bibinfo{year}{2020}\natexlab{}.
\newblock \showarticletitle{Attention-Based Activation Pruning to Reduce Data
  Movement in Real-Time AI: A Case-Study on Local Motion Planning in Autonomous
  Vehicles}.
\newblock \bibinfo{journal}{\emph{IEEE Journal on Emerging and Selected Topics
  in Circuits and Systems}} \bibinfo{volume}{10}, \bibinfo{number}{3}
  (\bibinfo{year}{2020}), \bibinfo{pages}{306--319}.
\newblock


\bibitem[\protect\citeauthoryear{Sen}{Sen}{2016}]%
        {sen2016context}
\bibfield{author}{\bibinfo{person}{Shreyas Sen}.}
  \bibinfo{year}{2016}\natexlab{}.
\newblock \showarticletitle{Context-aware energy-efficient communication for
  IoT sensor nodes}. In \bibinfo{booktitle}{\emph{DAC '16}}. IEEE,
  \bibinfo{pages}{1--6}.
\newblock


\bibitem[\protect\citeauthoryear{Sheeny~\textit{et al.}}{Sheeny~\textit{et
  al.}}{2020}]%
        {sheeny2020radiate}
\bibfield{author}{\bibinfo{person}{Marcel Sheeny~\textit{et al.}}}
  \bibinfo{year}{2020}\natexlab{}.
\newblock \showarticletitle{{RADIATE}: A radar dataset for automotive
  perception}.
\newblock \bibinfo{journal}{\emph{arXiv preprint arXiv:2010.09076}}
  \bibinfo{volume}{3}, \bibinfo{number}{4} (\bibinfo{year}{2020}).
\newblock


\bibitem[\protect\citeauthoryear{Solovyev~\textit{et al.}}{Solovyev~\textit{et
  al.}}{2021}]%
        {solovyev2021weighted}
\bibfield{author}{\bibinfo{person}{Roman Solovyev~\textit{et al.}}}
  \bibinfo{year}{2021}\natexlab{}.
\newblock \showarticletitle{Weighted boxes fusion: Ensembling boxes from
  different object detection models}.
\newblock \bibinfo{journal}{\emph{Image and Vision Computing}}
  \bibinfo{volume}{107} (\bibinfo{year}{2021}), \bibinfo{pages}{104117}.
\newblock


\bibitem[\protect\citeauthoryear{Stereolabs}{Stereolabs}{2021}]%
        {zeddatasheet}
\bibfield{author}{\bibinfo{person}{Stereolabs}.}
  \bibinfo{year}{2021}\natexlab{}.
\newblock \bibinfo{title}{{ZED Camera and SDK Overview}}.
\newblock
\newblock
\urldef\tempurl%
\url{https://cdn.stereolabs.com/assets/datasheets/zed-camera-datasheet.pdf}
\showURL{%
\tempurl}


\bibitem[\protect\citeauthoryear{Tann~\textit{et al.}}{Tann~\textit{et
  al.}}{2016}]%
        {tann2016runtime}
\bibfield{author}{\bibinfo{person}{Hokchhay Tann~\textit{et al.}}}
  \bibinfo{year}{2016}\natexlab{}.
\newblock \showarticletitle{Runtime configurable deep neural networks for
  energy-accuracy trade-off}. In \bibinfo{booktitle}{\emph{2016 CODES + ISSS}}.
  IEEE, \bibinfo{pages}{1--10}.
\newblock


\bibitem[\protect\citeauthoryear{Vatanparvar~\textit{et
  al.}}{Vatanparvar~\textit{et al.}}{2015}]%
        {vatanparvar2015battery}
\bibfield{author}{\bibinfo{person}{Korosh Vatanparvar~\textit{et al.}}}
  \bibinfo{year}{2015}\natexlab{}.
\newblock \showarticletitle{Battery lifetime-aware automotive climate control
  for electric vehicles}. In \bibinfo{booktitle}{\emph{DAC '15}}. IEEE,
  \bibinfo{pages}{1--6}.
\newblock


\bibitem[\protect\citeauthoryear{Zhang~\textit{et al.}}{Zhang~\textit{et
  al.}}{2018}]%
        {zhang2018exploring}
\bibfield{author}{\bibinfo{person}{Boyu Zhang~\textit{et al.}}}
  \bibinfo{year}{2018}\natexlab{}.
\newblock \showarticletitle{Exploring energy and accuracy tradeoff in structure
  simplification of trained deep neural networks}.
\newblock \bibinfo{journal}{\emph{IEEE Journal on Emerging and Selected Topics
  in Circuits and Systems}} \bibinfo{volume}{8}, \bibinfo{number}{4}
  (\bibinfo{year}{2018}), \bibinfo{pages}{836--848}.
\newblock


\end{thebibliography}


\end{document}